# Hope Speech detection in under-resourced Kannada language

**Adeep Hande** · **Ruba Priyadharshini** ·
**Anbukkarasi Sampath** ·
**Kingston Pal Thamburaj Prabakaran
Chandran** ·
**Bharathi Raja Chakravarthi**

**Abstract** Numerous methods have been developed to monitor the spread of neg- ativity in modern years by eliminating vulgar, offensive, and fierce comments from social media platforms. However, there are relatively lesser amounts of study that converges on embracing positivity, reinforcing supportive and reassuring content in online forums. Consequently, we propose creating an English-Kannada Hope speech dataset, KanHope and comparing several experiments to benchmark the dataset. The dataset consists of 6,176 user-generated comments in code mixed Kannada scraped from YouTube and manually annotated as bearing hope speech or Not- hope speech. In addition, we introduce DC-BERT4HOPE, a dual-channel model that uses the English translation of KanHope for additional training to promote hope speech detection. The approach achieves a weighted F1-score of 0.756, bet- tering other models. Henceforth, KanHope aims to instigate research in Kannada while broadly promoting researchers to take a pragmatic approach towards online

Adeep Hande
Indian Institute of Information Technology Tiruchirappalli, Tamil Nadu, India
*adeeph18c@iiitt.ac.in*

Ruba Priyadharshini
ULTRA Arts and Science College, Madurai Kamaraj University, Madurai, Tamil Nadu, India
*rubapriyadharshini.a@gmail.com*

Anbukkarasi Sampath
Kongu Engineering College, Erode, Tamil Nadu, India
*anbu.1318@gmail.com*

Kingston Pal Thamburaj
Sultan Idris Education University, Tanjong Malim, Perak, Malaysia
*fkingston@gmail.com*

Prabakaran Chandran
Mu Sigma Inc., Bengaluru, Karnataka, India
*prabakaran.chandran98@gmail.com*

Bharathi Raja Chakravarthi*
Insight SFI Research Centre for Data Analytics, Data Science Institute, National University of Ireland Galway, Galway, Ireland
*bharathi.raja@insight-centre.org*

content that encourages, positive, and supportive. We have published the data[1] and the corresponding codes[2] to support our claims.

**Keywords** Hope Speech · Code-mixing · Under-resourced languages

# 1 Introduction

The past decade has witnessed tremendous growth in social media users, mainly due to more comfortable access to the internet due to the modernization of coun- tries worldwide [1]. The surge has also resulted in several minority groups seeking support and reassurance on social media. The ongoing pandemic has led people to spend more time in their lives on social media to socialize despite social distancing norms [2, 3]. However, this poses a severe threat to adolescents and young adults who are ardent internet users. Social media applications such as Facebook, Twitter, YouTube have become an integral part of their daily lives [4]. While these plat- forms are a boon for youngsters as they can socialize more with others, they can also be a bane, which could be a significant factor for mental health problems [5], primarily due to the absence of content moderation on social media, which often entails offensive, abusive, misleading towards a particular group; usually, a minority [6]. Certain ethnic groups or individuals fall victim to manipulating social media to foster destructive or disruptive behaviour, a common scenario in cyberbullying [7, 8]. There have been several recent developments for hate speech and offensive language detection [9]. However, these systems disregard the potential biases present in the dataset that they are trained on and could hurt a specific group of social media users, often leading to gender/racial discrimination among its users [10,11,12].

Consequentially, there is a need to detect hope speech among social media. As Equality, Diversity, and Inclusion is an important topic as it also emphasizes the inclusion of a wide variety of people. We define *hope* as a form of reliance that the existing circumstances will change for the better [13]. Several Marginalized groups seek comfort and aid from content on social media that they can feel relatable to and can empathize with others' conditions [14]. These groups usually are people of marginalized communities, such as Lesbian, Gay, Bisexual, Transgender, Intersex, and Queer, Questioning (LGBTIQ) communities, racial and gender minorities. They perceive social media as one of the sources of counselling services, thus improving their emotional states [15, 14]. This form of speech is vital to everyone as they encour- age to improve the quality of life by taking action towards it. Hope speech aims to inspire people battling depression, loneliness, and stress by assuring promise, reassur- ance, suggestions, and support [16]. As most of the social media still revolve around English in multilingual communities, the phenomenon of code-mixing is prevalent in them. Studies have shown that code-mixing is an integral part of social media in multilingual countries [17, 18]. Code-mixing is the phenomenon of interchangeability between two or more languages during a conversation [19]. We observe code-mixing in our corpus, which represents the intrasentential modifications of codes. However, owing to the limited resources available in Kannada-English code-mixed text, our primary focus remains on constructing the corpus and conducting experiments to serve as a benchmark. Our dataset is distinct from HopeEDI [14], as that dataset spanned over three languages, namely, English, Tamil, and Malayalam, while our dataset focuses more on the dataset construction in code-mixed Kannada-English. While HopeEDI consisted of three classes: *Hope*, *Not-hope*, and *Other language*, our dataset consist consists of two classes: *Hope* and *Not-Hope*.

---

[1] https://zenodo.org/record/5006517/

[2] https://github.com/adeepH/kan_hope

Hence, we introduce KanHope, an English-Kannada code-mixed Hope Speech dataset aiming to minimize the scarcity in data availability for detecting hope speech in Kannada.

The principal contributions of the paper are listed below:

1. We have created the first dataset in Kannada to detect hope speech in code-mixed Kannada, to alleviate mental health problems on social media.
2. We provide a strong benchmark for the Kannada-English Hope speech dataset.
3. We propose DC-BERT4HOPE, a dual-channel language model based on the architecture of BERT that uses the translation of the dataset as additional input for training, performing better in contrast to the typical fine-tuned multilingual BERT.
4. We perform a comprehensive analysis of our models on the dataset along with a thorough error analysis on its predictions on the dataset.

## 2 Related Work

There has been significant research on extracting data from social media, especially exploiting user comments on YouTube, Facebook, and Twitter [20, 21, 22]. Most of the information extracted from social media do not follow any grammatical rules and tend to be written in code-mixed, or non-native scripts, which is generally observed among users from a multilingual country [18, 19, ?]. As people use social media plat- forms to educate themselves about current affairs, the users' comments are highly correlated with the events taking place throughout the world. For other under- resourced languages, researchers constructed corpora that were manually annotated for two tasks, namely, sentiment analysis and offensive language detection, in Tamil and Malayalam, consisting of 6,739 and 15,744 comments [20, 23]. To improve the research in this domain, shared tasks were conducted for sentiment analysis [24], and offensive language detection in Dravidian languages [25]. Many researchers have made efforts to detect offensive language. People can communicate without face- to-face interaction on social media, and they are susceptible to misunderstandings as they do not consider others' perspectives. Offensive speech is often used among social media forums to dictate others [26, 27]. Several deep learning frameworks were developed to classify hate speech into racist, sexist, or neither [28]. For code mixed languages, researchers created datasets for detecting hate speech in code mixed Hindi [29, 30]. However, there is a scarcity of data entailing hope speech detection. Previ- ously, very few works on hope speech detection, with the only dataset contribution being a sizeable multilingual corpus manually annotated for English, Tamil, and Malayalam, consisting of around 28K, 20K, and 10K comments, respectively [14]. Several other methods to alleviate gender/racial bias in Natural Language process- ing have been extensively studied for English [31], And in neural machine translation in French [32], for equality and diversity.

Several researchers have worked on engendering positivity on social media by developing and analysis of systems that filter out malignancy on social media by fo- cusing on very specific events such as crisis and war [33], inter-country social media dynamics [34, 35, 36], protests [37]. Other researchers tried to harness code-switching to sample hope speech and used it along with an English language identifier to re- treive texts in Romanized Hindi [38]. A classifier was developed using active learning strategies to support the minority Rohingya community [39]. A study involving a curated analysis of corpus of the movie industry to identify potential biases towards gender, skin colour, and gender representation was carried out [40, 41]. To encourage more research into hope speech for English, Malayalam, and Tamil, the authors con- ducted a shared task on hope speech detection for comments scraped from YouTube in these languages [42]. The organizers for the shared task used the Multilingual hope speech dataset, HopeEDI [14]. The corpus consists of 28,451 sentences in English, while 20,198 sentences in Tamil and 10,705 sentences in Malayalam. The authors of HopeEDI had set the baselines using preliminary machine learning algorithms yield- ing a weighted F1-score of 0.90, 0.56, and 0.73 for English, Tamil, and Malayalam, respectively, for their test sets in the shared task. The shared task on Hope speech detection saw a total of 30, 31, and 30 submissions for the final testing phase for the three languages, in the same order as mentioned above. Fine-tuning a pretrained XLM-RoBERTa model achieved the best-weighted F1-score of 0.854 in Malayalam [43]. An ensemble of synthetically generating code-mixed data for training ULMFiT, baseline-KNN, and a fine-tuned RoBERTa achieved the best score of 0.61 in Tamil [44]. For English, the authors combined the pretrained XLM-RoBERTa language model and the Tf-Idf vectors and fed them as inputs to an inception block which achieved a score of 0.93 [45].

Even though there has been researching aplenty on extracting data from social media, the same is not valid for code-mixed Kannada. Several corpora were created and were manually annotated for tasks of sentiment analysis and offensive language detection [46], And for emotion prediction [47], At the same time, several worked on developing models on sentiment analysis in code mixed Kannada [48, 49]. For devel- oping Language Identification systems (LID) in code-mixed languages, researchers constructed a Kannada-English dataset [50]. A stance detection system had been developed using sentence embeddings for code-mixed Kannada [51]. A second-order Hidden Markov Model (HMM) and Conditional Random Fields (CRF) are among the probabilistic classifiers used for Part-of-Speech (POS) tagging of Kannada lan- guage [52]. Regarding hope speech detection, we believe the corpus we created is the first Kannada-English code-mixed corpus.

## 3 Kannada

Kannada (ISO 639-3:kan) is one of the low-resourced Dravidian languages of India. Dravidian languages belong to a language family spoken by over 200 million peo- ple, predominantly in southern India and northern Sri Lanka [53, 54]. Despite its abundance in terms of speakers, Dravidian Languages are of low resource concerning language technology [55]. The language is primarily spoken by people in Karnataka, India, and is also recognized as an official language of the state [46]. The Kannada script, also called Catanese, is an alphasyllabary of the scripts of the Brahmic family evolving into the Kadamba script [56]. While Kannada is an under-resourced Dravid- ian language, its scripts write other under-resourced languages like Tulu, Konkani, and Sankethi [57].  The Kannada script has 13 vowels  (14 if the obsolete vowel   is included), 34 consonants, and 2 yogavahakas (semiconsonants: part-vowel,part-consonant) [58, 57]. The Kannada language has over 43 Million[3] speakers. However, as stated earlier, the lack of language technologies in Kannada makes it an under- resourced language.

## 4 Dataset Construction

People tend to express their opinions about many things on YouTube[4]. Due to its wide user base in India (more than 265 Million)[5] and being a multilingual country, we were motivated to extract the comments to work on code-mixed texts. The com- ments are collected using YouTube Comment Scraper[6]. We gather comments from several videos on distinctive topics such as movie trailers, India-China border dis- pute, people's opinion about the ban on several mobile apps in India, Mahabharata, and other social issues that involved oppression, marginalization, and mental health. Certain keywords are used to discover the videos and later used the scraping tool to extract the comments. We constructed the dataset between February 2020 and August 2020. The dataset is available on Huggingface datasets[7] [59].

Fig 1 represents the steps undertaken to construct the datasets and develop the models for our dataset. Our first step was to collect the videos from YouTube, followed by scraping them. We preprocess the scraped comments as discussed in Section 4.5, and we annotate the preprocessed texts using Google Forms. Each form consists of about 100 sentences, and the annotator has selected the label accordingly, as shown in Fig 2. These annotations are combined to form the dataset. The dataset initially consisted of three labels, *Hope*, *Not-Hope*, and *Not-Kannada*. We have deleted the comments that contain the *Not-kannada* label, as they do not contribute much to the overall development of the dataset. After constructing the dataset, we train the dataset to several machine learning algorithms and language models. We evaluate the models' predictions on the test set using Precision, Recall, and F1-Score metrics.

### 4.1 Hope Speech

For a person, *Hope* can be defined as an inspiration to people battling depression, loneliness, and stress by assuring promise, reassurance, suggestions, and support [14]. Hope can either be perceived as an emotion or as a theory [60]. Hope can also be defined as a perception to develop pathways to desired goals while self-motivating oneself into thinking to use those pathways [13]. The perception of Hope varies with age group as adults may lie on the high-hope scale, while children lie on the low-hope scale [13, 61]. Most studies assume children are goal-oriented; thus, their thoughts are related to sustainable action to achieve their goals [62]. Hope speech incites an optimistic perception of people's goals/aspects of life while also being vulnerable to negative influences [63]. Taking several definitions of Hope into consideration, we resort to a broader definition of hope speech. We define hope speech for our purpose

---

[3] https://www.ethnologue.com/language/kan

[4] https://www.youtube.com/

[5] https://www.omnicoreagency.com/youtube-statistics/

[6] https://github.com/philbot9/youtube-comment-scraper-cli

[7] https://huggingface.co/datasets/kan_hope

**Figure 1** Steps involved in the Kannada Hope speech dataset construction and modeling.

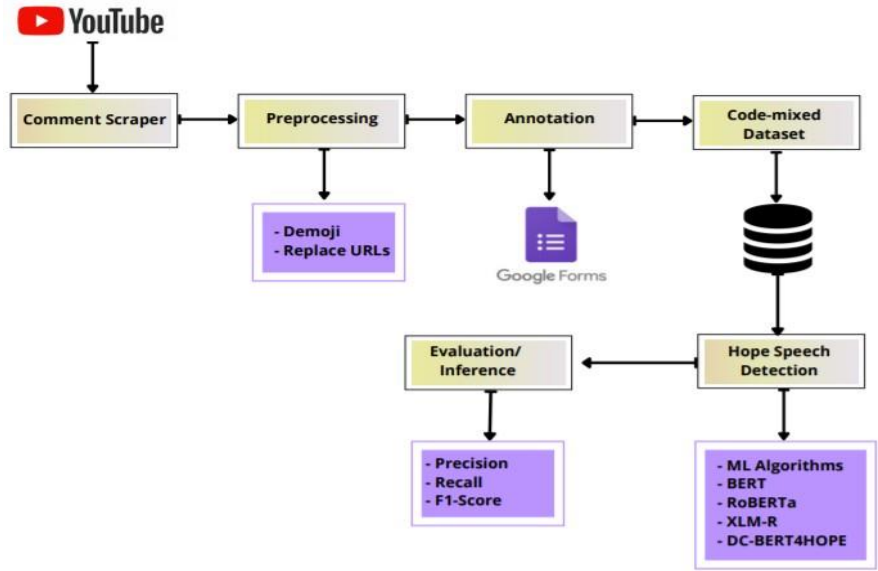

as "Any comments/texts that extend reassurance, aspiration, desires, support, or optimism to a person."

In the era of monetising online content, social media influencers/content creators often play an essential role in developing people's perception towards any entity, be it a brand or a socially sensitive topic, as they usually take a stance towards it [64, 65]. This practice can also harm the mental health of other users, often feeling marginalised, unnerved, or scared [66]. Moderating such target-specific content on social media would be ideal for the better mental health of its users. Our work aims to alter the conventional reasoning method by opting for a supportive, trustworthy, and righteous quality based on YouTube users' comments. Thus, we have provided instructions to the annotators to label them based on the followingguidelines:

**Hope speech:**
- Hope can be defined as an optimised state of mind that relies on a desire for positive results in the occasions and conditions of one's life or the world at large and can be present or future-oriented.
- Hope roots from inspirational talks about how people face gruelling situations and overcome them.
- Hope speech engenders cheerfulness and resilience that may positively impact several aspects of life, including work.
- The comment comprises an inspiration provided to participants by their peers and others, offering reassurance and insight.
- Comment talks about equality, diversity, and inclusion
- Comment talks about the survival story of people from marginalised groups.

**Non-hope speech**

- The comment produces hatred towards a person or a marginalised group.
- The comment is very discriminatory and attacks people without thinking of the consequences.
- The comment comprises racially, ethnically, sexually, or nationally motivated slurs.
- The comments do not inspire Hope in the readers' mind.
- The comment actively seeks violence and is reprimanding in nature.
- The comment is biased towards a product, taking any consequences into account, for the people who work in the company/organisation.

Some examples of Hope speech and Not-hope speech classes are:

- $T_1$: ತುಂಬು ಹೃದಯದ ಶುಭಾಶಯಗಳು ಕನ್ನಡ ಚಿತರಂಗದ ಅಭಿಮಾನಿಗಳಿಂದ
  **Transliteration:** Tumbu hrdayada śubhāśayagalu kannada citrarangada abhimanigalinda.
  **Translation:** Best wishes to the Kannada Cinema Industy from the bottom of my heart.
  **Label: Hope**
  This comment is classified as hope, as the speaker motivates and inspires the reader by his/her/their greetings to the Kannada Cinema Industry; Hence the comment instigates hope to its readers.

- $T_2$: ಸಾರ್ ನಿಮ್ಮ ತಂದೆ ನಿಮಗೆ ಕಲಿಸಿದ ಸಂಸಾಕಾರ ಸಂಸ್ಕೃತಿ ನಮಗೆ ತುಂಬಾ ಇಷ್ಟ ಆಯು ಮತ್ತು ನೀವು ಅವರು ತೋರಿಸಿದ ಮಾರ್ಗದರ್ಶನದಲಿಲ್ ನಡೀತಾ ಇರೋದು
  **Transliteration:** Sir nimma tande nimage kalisida sanskara sansthe namage thumba ishta aytu mattu neevu avaru toresida marghadharshanadalli nadita erodu
  **Translation:** Sir I like the culture your father had taught you, I hope you follow the path he guides you in.
  **Label: Hope**
  The sentence is classified as hope, due to the nature of the comment, appreciating the cultures and the behavioural knowledge interpreted by the son from his father.

- $T_3$: Yaru tension agbede yakandre dislike madiravru mindrika kadeyavru
  **Translation:** No one needs to worry as the people who disliked this are fans of Mandrika
  **Label: Not-hope**
  This sentence is classified as Not-hope. Despite the comment consoling someone because their opinion was disliked, the comment spreads hate to the person named Mandrika.

- $T_4$: ಟೊರ್ಲ್ ಅಂದರ್, ಬೊರ್ ನಾನಂ ಟಿಕ್ ಟಾಕಕೆ ಗಅಡಿಕ್ಟ್ ಆಗಿದ ಬಟ್ ನಮ್ ದೇಶಕ್ಕಿಂತ ದೊಡ್ಡು, ಈ ಟಿಕ್ ಟಾಕ್ಅಪೆಟ್ ಇನೊನ್ ಇಂದ್ ವಿಷಯ್ ನಮ್ ದೇಶದ್ ರೊಫೊಸೊ ಡೌನೊಲ್ಂಡ್ ಮಾಡಿ ಒಪನ್ ಮಾಡಿ ನೊಡುದಂ
  **Transliteration:** Troll andre, bro naanu tiktok ge addict agide but namma deshakkinta doddadalla, ee tiktok ashte ennond namm deshada rofoso download maadi nodu.
  **Translation:** For Troll, bro, I am addicted to TikTok, but it is not bigger than our nation; download our own Indian app Rofoso.
  **Label: Not-hope**
  This comment can be classified as Not-hope. Even though the comment states that TikTok is not more significant than the nation, expressing patriotism, the comment may or may not be factually correct. Hence, the comment spews unnecessary hatred towards TikTok.

## 4.2 Code-Mixing

Code-Mixing is often referred to as coupling linguistic units from two or more lan- guages into a conversation, or text [17, 27]. This phenomenon is prominent in speak- ers in multilingual countries [18]. Research has shown that code-mixing is more frequent in multilingual countries and is independent of illiteracy/inadequate knowl- edge [67, 19]. Kannada is a morphologically rich language [46]. We observe six different types of code-mixing in the dataset [20, 23]. A detailed description of the different types of code-mixing is shown below.

- Type 1: No code-mixing.
  $S_1$: ತುಂಬಾ ವಷರ್ದಾ ಹಿಂದೆ ಕೇಳಿದೆದ್. ಈಗ ಸಿಕಿಕ್ ದು ಸಕತ್ ಖುಷಿ ಆ
  **Transliteration:** Thumba varshada hinde kelidde, eega sikkiddu sakat khushi aa
  **Translation:** "Had heard so many years ago. Very happy that I Got it now" This sentence does not have any code-mixing and is written in a single language (Kannada).

- Type 2: Inter-sentential mix
  $S_2$: **Sister** ಹಾಗೆಲಾಲ್ ಮಾಡಲಲಾಲ್ ನಾವು ಯಾರಾದರೂ ತಪುಪ್ ಮಾಡಿತ್ಿದ್ೕರೆ ಅಂದಾಗ ಅವರನುನ್ *roast* ಮಾಡಿತ್ವ್ಿ ಅಷೆಟ್ೕ. ಅದು *entertainment* ಗೆ ಅಷೆಟ್ೕ. ತುಯಾಧನೆಯ್ಾದಗಳು ಹೀಗೆ *support* ಮಾಡಿತ್ೕರಿ
  **Transliteration:** Sister hagella madalla naavu yaradru tappu maadtidderi andaga avarannu roast ]madteve ashte. Adu entertainment ge ashte, thumba dhanyavadagalu heege support maadteri
  **Translation:** "We dont do that here sister, we only roast people if they commit some mistakes. It is solely for entertainment purposes, not to hurt others feelings. Keep supporting us"
  This sentence can be classified as Inter-sentential code-mixing, as there is a mix of English and Kannada, while Kannada is written in Kannada.

- Type 3: Only Kannada (Written in Latin Script)
  $S_3$: *Namma deshanu china thara aitu andre badatanane erala sir* **Transliteration:** Namma deshanu china thara aitu andre badatanane erala sir **Translation:** "If our country becomes like China, poverty ceases to exist, sir" This sentence is solely in Kannada, written in Latin Script.

- Type 4: Code-mixing at a morphological level
  $S_4$: ಈ ದರಿದ ಬಡತೆಲ್ಯ್ ತವ *tiktok* ನೂಡಿಯನಮಮ್ಮೋದಿಯವರು ಬಾಯ್ ಮಾಡಿದು ಗುರು
  **Transliteration:** Ee daridra bidtiva tiktok nodiya namma modiyavaru byan madiddu guru
  **Translation:** "After having a look at this dreadful app, tiktok, Modi banned it mate." This sentence can be classified as code-mixing at a morphological level, as the texts are written in both Kannada and Latin scripts.

- Type 5: Intra-sentential mixing $S_5$: *Estella matadonu elli matadodkintta border ge hogi matado maraya*
  **Transliteration:** Estella matdonu elli matadokintta border ge hogi matado maraya
  **Translation:** "If your only intention is to babble, do the same near the border, mate. "
  This sentence can be categorized as an Intra-sentential mix of English and Kannada and is only written in Latin.

- Type 6: Inter-sentential and intrasentential mix.
  This sentence can be sorted as an Inter-sentential and intrasentential mix, with Kannada being written in Latin and Kannada scripts.
  $S_6$: ನಿಜವಾಗಿಯೂ ಅದುತೆ *hartly heltidini... plz avrigella namma nimmellara support beku*
  **Transliteration:** Nijavahiyu adbutha hartly heltidini.. plz avrigella namma nimmellara support beku
  **Translation:** "Truly remarkable, saying it from the bottom of my heart, please, all of us need to support them"

### 4.3 Annotators

Due to the vast options available, we use Google Forms to collect annotations from the volunteers. The annotators' background information is collected to be aware of the diverseness among them. As described in Table 2, most of the volunteers' medium of schooling is in English. At the same time, all of their native languages are Kannada, as all hail from Karnataka, India. A minimum of three annotators annotated each form. The annotators adhere to the guidelines set forth by us, as discussed in Section 4.1. The annotation quality is validated using inter-annotator agreement and Krippendorff's alpha ($\alpha$) [68, 69], as a measure of the characteristics of the annotation setup. This statistical measure can compute reliability if the data is incomplete. Thus, annotators need not annotate every sentence. Using the nominal metric, we produced an agreement of 0.75 for the annotations.

### 4.4 Ambiguous Comments

In this section, we present some ambiguous comments that the annotators found it difficult to annotate. We asked the annotators to report to us the comments they found hard to annotate. We consider a given comment as an ambiguous one if more than one annotator has reported the comment. The red color for the english words, while blue for the transliterated kannada words:

- $A_1$: Sir ನಮ್ಮ್ ಲಂ communist ಆಡಳಿತ ಬಂದರೆ ಯಾವ ರೀತಿ ಇರಬಹುದು ಎಂಬುವುದರ ಮಾಹಿತಿ ಬೇಕು ಮತ್ತ್ ಅನುಕೂಲ ಅನಾನುಕೂಲ ಗಳ ಬಗೆ ತಿಳಿಸಿ ಕೊಡಿ ದಯವಿಟು
  **Transliteration:** Sir, nammallu communist radalita bandare yaav reeti irabahudu embuvudara mahiti beku mattu anukoola ananukula gala bagge thilisi kooda

**Figure 2** An instance of the google Forms used for annotating the corpus.

**dayavittu.**

The sentence asks the other person to say the pros and cons of a communist governance, if at all that happens. This sentence was difficult for the annotators to comprehend and take a stance on Hope or Not-hope.

- $A_2$: **Nija Film ast channagidya nanu nodide nange kandita arta agle Illa**

  **Translation**: The film was so good that I never understood any part of it.

  While it may sound the user was stoked by the movie, the annotators informed us that the author could have also implied in a sarcastic tone.

- $A_3$: ಈ ಚಿತ್ರದ ಲ್ಲಿ ಯಾವಲ್ ರೇಂಜ್ ಗೆ ಗ್ರಾಫಿಕ್ಸ್ ಮಾಡಿದಾದ್ರೆ ಅಂತ ಮತೊಬ್ಬ vfx ಆಟಿಸ್ಟ್ ಗೆ ಗೊತ್ತಾತ್ಗೋದು.

  **Transliteration:** **Ee chitradalli yaav range ge graphics madiddare antha mattoba vfx artist ge gottagodu.**

  **Translation**: The variety of graphics used in this movie can only be understood by a VFX graphics artist. As stated earlier, this statement could well have been implied in a sarcastic tone.

- $A_4$: Nanna ekkada ನನನ್ ಎಕಕ್ ಡ

  **Transliteration:** **Nanna ekkada nanna ekkada.**

  **Translation:** Where is your daddy? Where is he?

  This comment is an ambiguous one as its a *Telugu* sentence written in Kannada and English, and has no meaning when written in Kannada.

4.5 Pre-Processing

As the data is extracted from the comments section of YouTube, preprocessing would be imperative. To better adapt algorithms to the dataset, we follow the steps for preprocessing comments as listed below.

1. URLs and other links are replaced by the word, URL.
2. The emojis are replaced by the words that the emoji represents, like happy, sad, among other emotions depicted by emojis. As emojis mainly depict a user's intention, it would be imperative to replace them with their meanings to pick up their cues. As most models are pretrained only on unlabelled text, we feel that it would be necessary.
3. Multiple spaces in a sentence and other special characters are removed as they do not contribute significantly to the overall intention.

## 5 Data Statistics and Analysis

After completing the annotations, the responses are then converted into a comma-separated value (.csv) file. They are merged into a single file leading to the Kannada- English hope speech dataset . We perform several distinctive experiments, including several machine learning and deep learning algorithms, to baseline the dataset for future work on hope speech detection in code-mixed Kannada.

| Language Pair | Kannada-English |
| --- | --- |
| Number of Tokens | 56,549 |
| Vocabulary Size | 18,807 |
| Number of Posts | 6,176 |
| Number of Sentences | 6,871 |
| Tokens per post | 9 |
| Sentences per post | 1 |

**Table 1** Dataset Statistics

The dataset consists of 6,176 comments and is the largest hope speech dataset in Kannada, as before this, there are no datasets in this domain. We use *nltk*[8] for tokenizing words and sentences and calculating the corpus statistics as shown in Table 3. We observe that the vocabulary size is enormous due to code-mixed data in a morphologically rich language.

Table 2 represents the class-wise distribution of our dataset, along with the splits during training. We observe that Non-hope speech comprises the major portions of the dataset. Post annotation, the dataset comprised 7,572 comments with *Not- Kannada* which has a distribution of 1,396 out of 7,572 comments. The high number of other language label is a common scenario on information retrieved from user- generated content on online platforms. We have removed the comments labelled as *Not-Kannada*, resulting in the dataset consisting of 6,176 comments. The dataset is split into train, development, and test set. The training set comprises 80% of the distribution, while the development set consists of 10%, equal to the distribution of the test set. The class-wise distribution of data for the train, development and testing phase, as shown in Table 2. The classes are not equally distributed among the dataset, as Non-hope speech amasses 65.81%, while Hope speech spans the rest of it with a distribution of 34.19%. Fig 3 shows the difference in the distribution after removing the sentences having the *Not-Kannada* label. The dataset was evenly distributed with 54% for *Not-Hope*, 28% and 18% for *Hope* and *Not-Kannada*. How- ever, after removing the label, the class imbalance can be observed between the two datasets.

[8] https://www.nltk.org/

| Gender | Female Male | 3 |
| --- | --- | --- |
|  | Non-binary | 3 |
|  |  | 0 |
| Higher Education | Undergraduate | 5 |
|  | Graduate | 1 |
|  | Postgraduate | 0 |
| Medium of Schooling | English | 5 |
|  | Kannada | 1 |
| Total |  | 6 |

**Table 2** Annotators

**Figure 3** Comparison of labels distribution of the dataset before and after removing the third label.

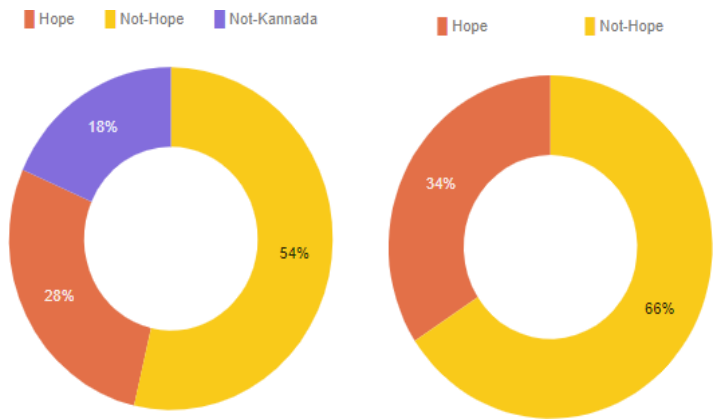

| Class | Non-hope Speech | Hope Speech | Total |
| --- | --- | --- | --- |
| Training | 3,265 | 1,675 | 4,940 |
| Development | 391 | 227 | 618 |
| Test | 408 | 210 | 618 |
| Total | 4,064 | 2,112 | 6,176 |

**Table 3** Class-wise distribution of Train-Development-Test Data

## 6 Methodology

We provide baselines for our dataset with a wide range of classifiers, from primitive machine learning algorithms to complex deep learning algorithms. We use the scikit- learn library [70] to tabulate our results. We perform our experiments as described below. To tabulate the results, we carried out an average of 5 runs on each model. As Kannada is a morphologically rich language, we refrain from using any stopwords or other lemmatisation approaches. We used the scikit-learn library for the machine learning algorithms. We used the Pytorch implementation of the pretrained lan- guage models available on Huggingface Transformers [9]. We fine-tuned the models on Google Colaboratory[10] for its easier access to GPU resources and User Interface.

6.1 Machine Learning Algorithms

*6.1.1 Logistic Regression*

Logistic Regression (LR) is a linear algorithm that uses the logistic function to model a binary dependent variable. LR was computed with L2 regularisation. The input features are the Term Frequency Inverse Document Frequency (TF-IDF) values of up to 5 grams, with the inverse regularisation parameter, C, set to 0.1. It is a control variable that retains the strength modification of regularisation by being inversely positioned to the lambda regulator.

*6.1.2 K-Nearest Neighbors*

K-Nearest Neighbors algorithm assumes the similarity between a new entity/data and available data to assess to put the entity/data into a category that is most similar to the categories available at hand. We used KNN for classification with 3, 4, 5, and 7 neighbours by applying uniform weights. We use *Minkowski* as the distance metric, with the power parameter (p) for the distance metric as 2 while setting uniform weights for the neighbours.

*6.1.3 Decision Tree and Random Forest*

A decision tree is a diagrammatic representation of classification, where the paths from the root to the leaf formalise the rules. The maximum depth was 800, and the minimum sample splits were 5, with *Gini* as the criterion. Random Forest is an ensembling method that is generally used for classification, regression tasks and operates by collecting many individual decision trees. The class with the maximum amount of predictions from decision trees is decided as the output class.

*6.1.4 Naive Bayes Classifier*

Naive Bayes is a probabilistic classifier that computes the probability of a hypothesis activity to a given evidence activity. It is based on Bayes' theorem with a 'naive' assumption that each pair of features are conditionally independent given the value of the output variable [70]. We evaluate a Naive Bayes classifier for multinomially distributed data, with $\alpha$ = 1 for Laplace smoothing to prevent the occurrence of zero probabilities.

---

[9] https://huggingface.co/transformers/
[10] https://colab.research.google.com/

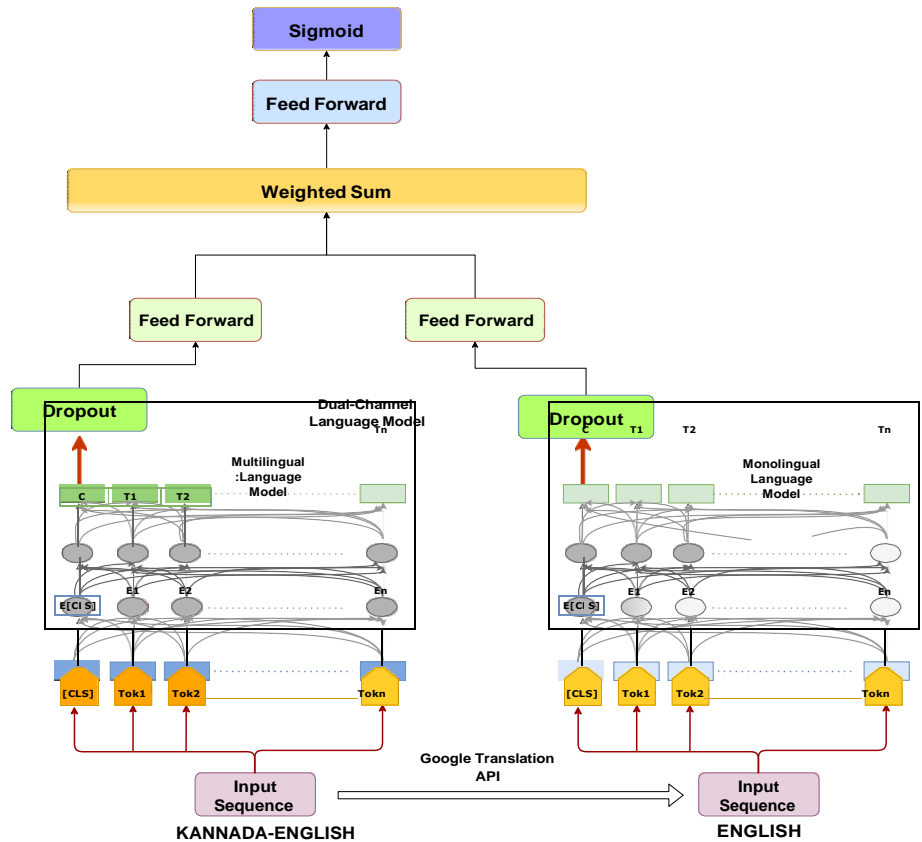

**Figure 4** Dual-Channel BERT-based Language Model [DC-BERT4HOPE]

6.2 Fine-tuning pretrained Language Models

The success of the transformer architecture [71] has led to the transfiguration of recurrent neural networks (RNN) to transformer-based models, resulting in language models adapting to transformers as their building blocks. For hope speech detection, we have fine-tuned four pretrained language models, all being based on the primary architecture of BERT. Since all models were pretrained on unlabelled monolingual or multilingual data, the models could face difficulties classifying code-mixed sentences. We use Binary Crossentropy as the loss function, as this is a binary classification task. We utilise Adam optimizer (AdamW) available on Huggingface Transformers by decoupling weight decay from the gradient update [72, 73]. The corpus is first tokenized to cleave the word into tokens. During tokenization, the special tokens needed for sentence classification, the [CLS] token at the start of a sentence and the [SEP] token at the end. Post the addition of the special tokens, the tokens are replaced by ids (*input_ids*), and *attention_masks* for training. During fine-tuning, we extract the pooled output of the [CLS] token and feed the output through an activation layer (Sigmoid) to compute the output prediction probabilities for the given sentence. The Sigmoid activation function is used for binary classification problems and is formulated as follows:

$$Sigmoid(x) = \frac{1}{1 + e^{-x}} \tag{1}$$

**Table 4** Hyper-parameters used for fine-tuning BERT-based language models

| Hyper-parameters | Characteristics |
|---|---|
| Optimizer | AdamW [72] |
| Batch Size | [32, 64, 128] |
| Dropout | 0.1 |
| Loss | Binary cross-entropy |
| Learning rate | 2e-5 |
| Max length | 128 |
| Epochs | 10 |

Where e is the Euler's number, and K is the number of classes. The hyperparameters we used for fine-tuning the pretrained language models are as shown in Table 4.

**Table 5** The pretrained models used for DC-BERT4HOPE (Dual-Channel BERT) Name

| | Pretrained Model |
|---|---|
| BERT | bert-base-uncased |
| mBERT | bert-base-multilingual-cased |
| XLM-R | xlm-roberta-base |
| RoBERTa | roberta-base |

### 6.2.1 BERT

We have used two language models that are pretrained architecture of BERT [74]. Unlike other unidirectional language models (GPT, ElMo), Bidirectional representa- tion from Transformers (BERT) gains from joint conditioning from both sides, left and right. BERT employs two pretraining strategies, *Masked Language Modeling* (MLM), where a specific portion of the unlabelled data (15%) are masked during pretraining, mainly believing the word spots themselves accidentally due to its na- ture of bidirectional representations. The other pretraining strategy is *Next Sentence Prediction* (NSP), indicating whether a given sentence follows the previous sentence. As shown in Table 5, we use **bert-base-uncased**, a monolingual language model that has been pretrained only on lower cased English text with a 12-layer, 768-hidden dimension, 12-heads, and 110 million parameters. Multilingual-BERT (mBERT) [75], a multilingual version of BERT, is pretrained on publicly available Wikipedia dumps of the top 100 languages. We use **bert-base-multilingual-cased**[11] which is pretrained on the cased text of the top 104 languages, with 12-layer, 768-hidden dimensions, 12-heads, and 179 million parameters. Both of the models follow the same parent architecture, however only differing on corpora used during pretraining.

*6.2.2 RoBERTa*

RoBERTa is a language model primarily based on BERT architecture, with only modifications on the hyperparameter optimization and better pretraining strategies [76]. Unlike BERT, RoBERTa disregards the Next Sentence Prediction (NSP) loss from its pretraining, as the authors did not find any improvement regardless of the loss function. During tokenization, RoBERTa uses a byte-pair encoding (BPE) rather than BERT's WordPiece tokenization. We use **robert-base**, a monolingual language model pretrained on 160GB of unlabelled English texts, with 12-layer, 768-hidden dimensions, 12-heads, and 125 million parameters.

*6.2.3 XLM-RoBERTa*

**Table 6** Class-wise Precision (P), Recall (R), and F1-Scores for both the classes of the dataset. DC-BERT4HOPE(model1-model2): model1: Monolingual, model2: Multilingual

| Model | Not-Hope | | | Hope | | | | | | |
|---|---|---|---|---|---|---|---|---|---|---|
|  | P | R | F1 | P | R | F1 | Acc | W(P) | W(R) | W(F1) |
| Logistic Regression | 0.681 | **0.964** | 0.798 | **0.788** | 0.228 | 0.354 | 0.693 | 0.721 | 0.693 | 0.634 |
| KNN | 0.705 | 0.890 | 0.787 | 0.659 | 0.364 | 0.469 | 0.696 | 0.688 | 0.696 | 0.670 |
| Decision Tree | 0.732 | 0.797 | 0.763 | 0.591 | 0.500 | 0.542 | 0.688 | 0.680 | 0.688 | 0.681 |
| Random Forest | 0.736 | 0.867 | 0.796 | 0.673 | 0.469 | 0.553 | 0.720 | 0.713 | 0.720 | 0.706 |
| Naive Bayes | 0.719 | 0.885 | 0.793 | 0.674 | 0.408 | 0.508 | 0.709 | 0.702 | 0.709 | 0.688 |
| mBERT | 0.757 | 0.854 | 0.802 | 0.680 | 0.531 | 0.596 | 0.735 | 0.728 | 0.735 | 0.726 |
| BERT | 0.758 | 0.780 | 0.769 | 0.604 | 0.575 | 0.589 | 0.704 | 0.701 | 0.704 | 0.702 |
| DC-BERT4HOPE(bert-mbert) | 0.771 | 0.836 | 0.802 | 0.672 | 0.575 | 0.619 | 0.740 | 0.734 | 0.740 | 0.735 |
| DC-BERT4HOPE(roberta-mbert) | **0.788** | 0.838 | **0.812** | 0.690 | 0.614 | **0.650** | **0.756** | **0.752** | **0.756** | **0.752** |
| DC-BERT4HOPE(roberta-xlm) | 0.777 | 0.779 | 0.778 | 0.621 | **0.618** | 0.620 | 0.720 | 0.720 | 0.720 | 0.720 |

XLM-RoBERTa relies on unsupervised cross-lingual learning at scale, implying that the language representations learnt from one language would benefit the other, indicating that the model would improve the performance on code-mixed data. We use **xlm-robert-base**, the smaller version of the model, with 270 million parameters, 12-layers, 768-hidden-states, and 8-heads, while being trained on 2.5 TB of newly created clean CommonCrawl data in 100 languages.

*6.2.4 Dual-Channel BERT*

Inspired by the approach employed in MC-BERT4HATE [77], we propose a Dual- Channel BERT4Hope (DC-BERT4HOPE) by fine-tuning a language model based on BERT on the code-mixed data and its translation in English. For translating the code-mixed KanHope to English, we use the Googletrans API [12]. This API makes use of GoogleTrans Ajax API [13] to make calls to detect methods and translate. We call the *Translator* function and set the destination language to English, as the *Translator* attempts to identify the source of the language on its own. Using two channels of pretrained language models is dependent on the advancements of En- glish language models available. By translating the sentences to English, we have obtained additional training data for hope speech in English. We believe using Dual Channel BERT, one model for the code-mixed Kannada-English - a multilingual lan- guage model, while the other model for the translated English texts - a monolingual language model (pretrained on English) learn better from two languages instead of one. The weighted sum of the layer will take the weighted sum of two pooled out- puts obtained from the [CLS] token. We tokenized the code-mixed sentences with a pretrained multilingual tokenizer and the translated English sentences with a mono- lingual tokenizer pretrained on English for fine-tuning. The translated text was used as the input for the first channel (RoBERTa or BERT), while the standard raw text was fed as input to the multilingual language model (mBERT or XLM-RoBERTa). As shown in Fig 4, the pooled output was extracted from the [CLS] token of both the models, and a layer takes the weighted sum of both the pooled outputs. The overall output was then fed into a feed-forward network followed by a sigmoid activation function.

---
[11] https://github.com/google-research/bert/blob/master/multilingual.md

DC-BERT4HOPE(model1-model2) represents the dual-channel model that uses *model1* for the translated text while *model2* for the code-mixed texts. We use two language models based on BERT and RoBERTa for *model1* to train on the translated text. For the *model2*, we use two multilingual models, mBERT and XLM-RoBERTa. DC(bert-mbert): This model uses *bert-base-uncased* for the English text while *bert- base-multilingual-cased* language model for the code-mixed Kannada-English. The same approach is followed for every other Dual-Channel BERT models.

## 7 Results and Discussion

The experimental results of classifying hope speech in code mixed Kannada with various distinct techniques are shown in terms of precision and recall for the respec- tive classes, along with the overall accuracy, weighted averages of Precision, Recall, and F1-score, being tabulated in Table 6. The three metrics are computed as follows:

$$Precision(P) = \frac{TP}{TP + FP} \qquad (2)$$

$$Recall(R) = \frac{TP}{TP + FN} \qquad (3)$$

$$F1-Score = \frac{2*P*R}{P + R} \qquad (4)$$

TP, FP, and FN are True Positives, False Positives, and False Negatives. The weighted average takes the metrics from each class similar to the macro-average. However, the contribution of each class to the average is weighted by the number of examples available for it. The macro-average computes the metrics (precision, recall, F1-score) independently for each of the classes and then take the average of them, ignoring the presence of class imbalance, if any, the values of which are tabulated. We do not observe any major class imbalance in our dataset. Hence, we refrain from using the micro average, which aggregates the contributions of all classes to com- pute the average metric. For our test set, the number of samples on *not-hope speech* is 390, while 228 on *hope speech*. The codes of our experiments are available[14].

---

[12] https://pypi.org/project/googletrans/

[13] https://translate.google.com/

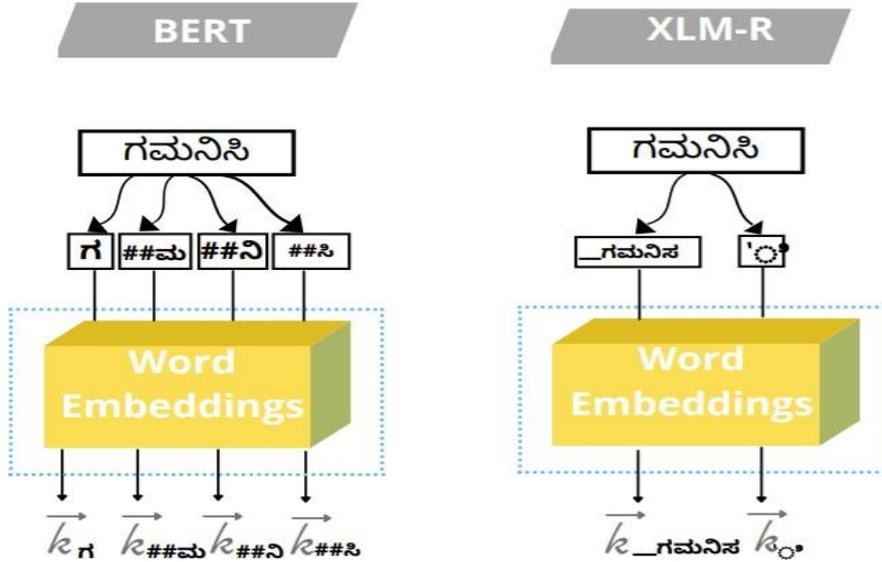

**Figure 5** WordPiece tokenization in BERT vs Byte-Pair Encoding in XLM-RoBERTa Word: *Gamanisi*, Translation: *Observe*.

From the tables, we observe that all the machine learning algorithms perform rea- sonably for the code mixed Kannada corpus. Multinomial Naive Bayes and Random Forest are among the machine learning classifiers that fared relatively better than other machine learning algorithms [78]. We observe that Decision trees and Logistic Regression are among the classifiers that perform poorer than the others. Due to the nature of logistic regression, where the algorithm estimates the linear boundary, indicating that the features are not significantly correlated to each other. The low correlation is also the reason why the Decision Tree classifier performs poorly. Due to the nature of the dataset, we used fine-tuned several pretrained language models. We use four language models for the dual-channel BERT4HOPE, listed in Table 5. We fine-tune multilingual BERT and the uncased base version of BERT separately to assess the significance of improving performance in DC-BERT4HOPE if any. Out of the two BERT models, multilingual BERT performs better than the BERT model that was pretrained only on English, with a minor increase of 2.1%. However, the performance between the machine learning algorithms and pretrained language mod-els differ by around 7.8%. We trained three dual-channel language models based on the possible combinations between the monolingual and multilingual models. *DC-BERT4HOPE (bert-mbert)* used the monolingual BERT (only English) for the trans- lated text, while the multilingual BERT for the code-mixed Kannada-English texts. DC-BERT4HOPE(bert-mbert) achieves a weighted F1-Score of 0.740, an improve- ment of 0.5% from mBERT and 3.6% from monolingual BERT. When *roberta-base* is used for the translated texts and multilingual BERT for the code-mixed texts, it achieves the best performance of all the models, having an F1-Score of 0.756 (+ 1.6% from the previous model). The principal reason for this increase comes down to the better hyper-parameter tuning and pretraining strategy used by RoBERTa, as multilingual BERT is used in both models, indicating that mBERT was not the reason for this increase in the performance.

[14] https://github.com/adeepH/KanHope

| Sentence | predictions | Real labels |
|---|---|---|
| Driver superb sir n Obbara hangilladanthe | Hope | Hope |
| Troll Stupid Fans war ge antha bandre nimmakka | Not-hope | Not-hope |
| ಕೇಳಿ ಕಾದಿರುವ ಭಾಂದವರೇ ಭುವಿಯಲಿಲ್ ಅವನ ಅರಿತವರೇ<br>keli kadiruva bhandavare n bhuviyalli avan arithavare n yara | Hope | Not-hope |
| Nandi Parthasarathi ille gotta aagta ide ninu | Not-hope | Not-hope |
| Unity of India ನಾವಂೂ ಬಹಮ್ ಣರಂ ಆಗುವುದೀಲ್ ಏಕಂದ<br>Unity of India naavu bhramanaru aguvudu ekanda | Not-hope | Not-hope |
| Awesome estu sari kellidru innod sari kellonn | Not-hope | Not-hope |
| ಸಾರಲ್ ನಿಮ್ಮಲ್ ತಂದೆ ನಿಮಗೆ ಕಲಿಸಿದ ಸಂಸಾಕ್ಕರ ಸಂಸ್ಕೃತಿ<br>sar nimma thande nimage kalisida sanskara sanskriti | Not-hope | Hope |
| ಪದಗಳು ಸೋತಿದೆ ಸಿನಿಮಾದ<br>padagalu sotide cinema da antargali ariyalu hradya | Hope | Hope |
| ಕಿಚಛ್ ನ ಹಾವಳಿ book my show Alli pailwan cr kuru<br>kicchana havali book my show Alli pailwan cr kuru | Not-hope | Not-hope |

**Table 7** Examples of predictions on the test set. *Preds* Indicates the predictions on the text while *real* represents the gold labels.

We have also trained DC-BERT4HOPE (roberta-xlmr), which uses *roberta-base* for the translated texts and *xlm-roberta-base* for the code-mixed texts. We observe that this model performs poorer than DC-BERT4HOPE (bert-mbert), despite XLM- RoBERTa being pretrained on 2.5 TB of data and its approach to an unsupervised cross-lingual learning scale. We believe that one of the reasons for the poor per- formance of XLM-R is its tokenizations. Even though the authors of XLM-R state that the performance of their model is independent of the types of encoding in tokenizations, it is found that Byte-Pair Encoding (BPE) tends to have a poorer morphological alignment with the original code-mixed text [79]. XLM-R uses BPE tokenizer in contrast to the WordPiece tokenization used in BERT, creating more subwords. The difference in the tokenization between mBERT and XLM-R is de- picted in Fig 5. However, as Kannada is a morphologically rich language, we believe XLM-RoBERTa performs poorly than multilingual BERT.

Fig 6 represents the confusion matrix for the test set on the best performing model. We observe that the model predict 327 out of 390 samples correctly for the *Not-hope* label, while the model predicts 140 out of 228 samples correctly for the other class. We observe that the precision, recall, and F1 scores are higher for *Not- hope Speech* when compared to the others. One of the main reasons could be the class-wise distribution among the dataset, as 4,064 sentences out of 6,176 belong to *Not-hope Speech*. To our surprise, we see that the monolingual BERT (only English) performed poorer than some machine learning algorithms, having poor precision, recall, and F1 scores. We believe that this roots in the characteristics of the dataset. The main objective of performing these experiments is to serve as a baseline for KanHope for researchers to develop more sophisticated models to broaden further research on positivity. We randomly chose nine sentences and their predictions and have listed them in Table 7. We observe that two sentences predict the wrong labels. The sentence **keli kadiruva bhandavare n bhuviyalli avan arithavare n yara** has been wrongfully classified as *Not-hope*. The sentence translates to **The morals and values that your father has taught you**. While this comment was identified as *Hope* by the annotators, the model predicts it otherwise; this could be due to the nature of the comment, as the sentence seems quite incomplete. The other sentence that the model identifies incorrectly is **keli kadiruva bhandavare n bhuviyalli avan arithavare n yara** as *hope*. As the sentence is an idiom that does not incite any hope, we believe that the model assumes idioms as inciting hopes mainly due to the nature of the idioms, as words in an idiom require commonsense knowledge has not been completely achieved by the pretrained models yet. Thus, some predictions have been wrong due to the lack of commonsense knowledge.

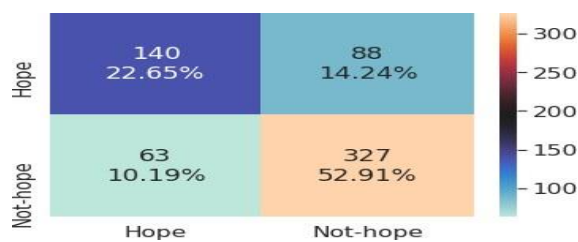

**Figure 6** Confusion matrix heatmap for DC-BERT4HOPE(RoBERTA + mBERT)

## 8 Conclusion

A surge in the active users on social media has inadvertently increased the amount of online content available on social media platforms. There is a need to motivate positivity and hope speech in platforms to instigate compassion and assert reassur- ance. In this paper, we have presented KanHope, a manually annotated code-mixed data of hope speech detection in an under-resourced language, Kannada, consisting of 6,176 comments crawled from YouTube. We also propose DC-BERT4HOPE, a Dual-Channel BERT-based model that uses the best of both worlds: Code-mixed Kannada-English and Translated English texts. Several pretrained multilingual and monolingual language models were analysed to find the best approach that yields a tremendous weighted F1-Score. We have also trained the dataset on preliminary machine learning algorithms to baseline for future work on the dataset. We believe that this dataset will expand further research into facilitating positivity and opti-mism on social media. We have developed several models to serve as a benchmark for this dataset. We aim to promote research in Kannada.

**Acknowledgements** The author Bharathi Raja Chakravarthi was supported in part by a research grant from Science Foundation Ireland (SFI) under Grant Number SFI/12/RC/2289_P2 (Insight_2), co-funded by the European Regional Development Fund and Irish Research Council grant IRCLA/2017/129 (CARDAMOM-Comparative Deep Models of Language for Minority and Historical Languages).

**Funding**

This research has not been funded by any company or organization

**Compliance with Ethical Standards**

**Conflict of interest:** The authors declare that they have no conflict of interest.

**Availability of data and material:** The dataset used in this pa- per are obtained from https://zenodo.org/record/4904729 and/or https://huggingface.co/datasets/kan_hope.

**Code availability:** The data and approaches discussed in this paper are available at https://github.com/adeepH/kan_hope.

**Ethical Approval:** This article does not contain any studies with human participants or animals performed by any of the authors.

KanHope 23

**Appendix**

We have listed the links to the videos here:

| Video Title | Links |
| --- | --- |
| "Sex: OTHER": A Kannada short film about transgenders | https://www.youtube.com/watch?v=eGhBPVG3DLo |
| ಡಿಂವರ್ ಅದಮ್ಲನೆ ಹಚು ಸಿನಿಮಂ ಮಾಡಿದು ..! | https://www.youtube.com/watch?v=Uudb9vK5n10 |
| ತೊಗರಿ ತಿಪಪ್ - ಭಾಗ ೧ | ಹಾಸಯ್ ನಾಟಕ | ಶಂಭು ಬಳಿಗಾರ ರವರ | | https://www.youtube.com/watch?v=vAkTwXpL7Vw |
| ಕಲಿಯುಗದಲಿಲ್ ಈಗ ಅಶವ್ ತಾಥ್ಯ ಎಲಿಲ್ಲ್ಾದ್ನೆ ಗೊತಾತ್? | https://www.youtube.com/watch?v=AHdCVZ8ws1M |
| KATHEYONDU HELUVE Engineering |#KannadaNew #ShortFilm | https://www.youtube.com/watch?v=_HoA1sI8z1A |
| ದಿಯಾಗೆ ಜನ ಫಿದಾ..! ಆದರೆ ಥೇಟರ್ ಗೆ ಬತಿರ್ಲ್ ..ಯಾಕೆ? | https://www.youtube.com/watch?v=mVpfXPGX-sY |
| Pogaru | Karabuu | Dhruva Sarja | Rashmika Mandanna | https://www.youtube.com/watch?v=Ysf4QRrcLGM |
| Halli College | Kannada Short Film| Avinash Chouhan | https://www.youtube.com/watch?v=-MJ7QkAVveI |
| Goudra Runa | Kannada Short Film | Indian short Film | https://www.youtube.com/watch?v=0Jog7Amz9_o |
| TIKTOK-RAJASREE|KIRIK GURU ROASTING | https://www.youtube.com/watch?v=_qkad1dWefk |
| ಡೊಂೀನ್ ಪತಾಪ್ ನ ಪಶಿಚ್ಛೆಳಟ್ ನಿಜಾನಾ? | https://www.youtube.com/watch?v=TMzsXk8VeeY |
| Gentleman | Kannada New Trailer 2020 | Prajwal Devaraj | https://www.youtube.com/watch?v=he4vdhjGUIc |
| ONDU SHIKARIYA KATHE | OFFICIAL TRAILER | https://www.youtube.com/watch?v=D7bxDmaIO3o |
| Avane Srimannarayana (Kannada) - Hands UP | https://www.youtube.com/watch?v=C3jlOlzSL8I |
| ಈಗ ಎಲಲ್ ರಿಗೂ ಭಾರತ-ಚೀನಾ ಆಧಿಕರ್ತೆ ಅಥರ್ ಆಗುತೆತ್ | https://www.youtube.com/watch?v=_fniGrfPqjM |
| How's life in CHINA now ? |Vlog 2| kannadiga | https://www.youtube.com/watch?v=ccjxoMt2fd0 |
| Tik Tok Ban Funny Roast kannada | Creative kannadiga | https://www.youtube.com/watch?v=K7lWuMHVW34 |
| Tik Tok Ban Kannada Roast | Ban Chinese Apps | boycott china | https://www.youtube.com/watch?v=G7iwocCFkyg |

**Table 8** The list of videos from which the comments we scraped